\pdfoutput=1

\documentclass[11pt]{article}

\usepackage{EMNLP2023}

\usepackage{subcaption}

\usepackage{times}
\usepackage{latexsym}

\usepackage[frozencache,cachedir=.]{minted}

\usepackage[T1]{fontenc}

\usepackage[utf8]{inputenc}

\usepackage{microtype}

\usepackage{inconsolata}

\usepackage{graphicx}
\usepackage{float}

\title{Learning the Bitter Lesson: Empirical Evidence from 20 Years of CVPR Proceedings}

\author{Mojtaba Yousefi\\
  Northeastern University / Boston, MA \\
  \texttt{yousefi.m@northeastern.edu} \\\And
  Jack Collins \\
  Develop Health / Menlo Park, CA \\
  \texttt{jack@develophealth.io} \\}

\begin{document}
\maketitle
\begin{abstract}
  This study examines the alignment of \emph{Conference on Computer Vision and Pattern Recognition} (CVPR) research with the principles of the "bitter lesson" proposed by Rich Sutton. We analyze two decades of CVPR abstracts and titles using large language models (LLMs) to assess the field's embracement of these principles. Our methodology leverages state-of-the-art natural language processing techniques to systematically evaluate the evolution of research approaches in computer vision. The results reveal significant trends in the adoption of general-purpose learning algorithms and the utilization of increased computational resources. We discuss the implications of these findings for the future direction of computer vision research and its potential impact on broader artificial intelligence development. This work contributes to the ongoing dialogue about the most effective strategies for advancing machine learning and computer vision, offering insights that may guide future research priorities and methodologies in the field.
\end{abstract}

\section{Introduction}

Rich Sutton's influential essay "The Bitter Lesson" argues that the most significant advancements in artificial intelligence (AI) have come from focusing on general methods that leverage computation rather than human-designed representations and knowledge. This principle has been particularly evident in the field of Computer Vision (CV), which has witnessed a notable shift from hand-crafted features to deep learning models.

In this paper, we investigate the extent to which the abstracts of the Conference on Computer Vision and Pattern Recognition (CVPR), a major machine learning (ML) conference, align with the principles of the "bitter lesson" over a span of 20 years. We analyze a random sample of 200 papers each year, addressing the following research questions:

\begin{itemize}
  \item How has the focus on general methods and computation evolved in CVPR abstracts over the past two decades?
  \item What trends can be observed in the adoption of deep learning techniques and the shift away from hand-engineered features?
  \item To what extent do the abstracts reflect the key insights of Sutton's "bitter lesson," and how has this alignment changed over time?
  \item Is there a significant relationship between a paper's alignment with the "bitter lesson" principles and its impact, as measured by citation count?
\end{itemize}

To address these questions, we employ large language models (LLMs), which themselves are a prime manifestation of principles outlined in the "bitter lesson", to analyze the CVPR abstracts. The evaluation is based on five metrics assigned by the LLMs, providing a comprehensive assessment of the alignment between the abstracts and the "bitter lesson."

Our research provides valuable insights into the overall direction of the ML community and reveals interesting trends in the adoption of Sutton's principles. By leveraging LLMs to analyze a large body of research literature, we offer a novel approach to understanding the learning and evolution of a scientific field. This method allows us to uncover patterns and trends that may not be immediately apparent through traditional research methods, providing a more comprehensive understanding of the current state of ML research and its alignment with the principles that have proven most effective in driving progress in AI.

The potential impact of our findings on future CV research directions is significant. By identifying trends in the adoption of general methods and deep learning techniques, we can inform the development of foundation models for CV at the state of the art. These insights contribute to a deeper understanding of the current state of ML research and highlight potential areas for further exploration and growth in the field.
\section{Background}

\subsection{The Bitter Lesson}
The field of artificial intelligence (AI) has witnessed a paradigm shift, eloquently articulated in Rich Sutton's influential essay "The Bitter Lesson" \citep{Sutton2019BitterLesson}. Sutton's thesis emphasizes the primacy of general methods that harness computational power over human-designed representations and domain-specific knowledge. This perspective echoes the seminal work of Leo Breiman, who, two decades earlier, delineated the dichotomy between statistical and algorithmic approaches in his paper "Statistical Modeling: The Two Cultures" \citep{Breiman2001StatisticalMT}. Breiman's insights, along with subsequent works like \citep{Halevy2009TheUE}, have profoundly shaped our understanding of data-driven methodologies in AI.

\subsection{Evolution of Computer Vision}

The field of Computer Vision (CV) exemplifies the principles of Sutton's "bitter lesson." Traditionally reliant on hand-crafted features like SIFT, HOG, and Haar cascades for object detection and image classification, CV underwent a paradigm shift with embracing deep learning, particularly Convolutional Neural Networks (CNNs). This transition enabled the automatic learning of hierarchical features directly from raw image data, eliminating the need for manual feature engineering and significantly improving performance across various CV tasks.

The 2012 ImageNet Large Scale Visual Recognition Challenge (ILSVRC) marked a pivotal moment in this evolution. AlexNet, a CNN architecture, achieved a remarkable 15.3\% top-5 error rate, outperforming previous models by over 10 percentage points. This breakthrough was facilitated by the convergence of ImageNet's massive annotated dataset, advancements in CNN architectures, GPU computing power, and foundational work of visionary researchers.

The subsequent emergence of foundation models further aligned CV with Sutton's principles. Models like CLIP, ALIGN, and Florence demonstrate remarkable adaptability across diverse tasks with minimal fine-tuning, leveraging extensive multimodal datasets to learn rich, transferable representations. For instance, the Florence model has achieved state-of-the-art results by integrating universal visual-language representations from web-scale image-text data \citep{10.1007/s00371-021-02166-7}.

This evolution from traditional feature engineering to deep learning and foundation models in CV underscores the importance of leveraging computation and vast datasets for superior performance and generalization. For a comprehensive overview of these advancements, readers may refer to \citet{Minaee2020ImageSU}, which details recent progress in deep learning for image segmentation.

\subsection{Large Language Models in Academic Evaluation}
The integration of Large Language Models (LLMs) into the evaluation of academic texts has emerged as a significant area of interest. LLMs, such as GPT-4, have demonstrated remarkable capabilities in processing and analyzing large volumes of information quickly, making them suitable for various applications, including the assessment of academic literature. For instance, research has shown that LLMs can effectively assist in title and abstract screening for literature reviews, which is crucial in the biomedical domain \citep{10.1186/s13643-024-02575-4}. Moreover, LLMs have been employed to perform qualitative data analysis, producing consistent results across multiple iterations \citep{10.1101/2023.07.17.549361}.

In addition to their analytical capabilities, LLMs have been shown to possess a degree of human-like judgment in evaluating the quality of text. The G-EVAL framework, which utilizes LLMs to assess the quality of natural language generation outputs, demonstrates that LLMs can align closely with human evaluators in certain contexts \citep{10.18653/v1/2023.emnlp-main.153}. However, the deployment of LLMs in academic evaluation is not without challenges. LLMs can exhibit biases similar to those found in human judgments, which may affect the fairness and accuracy of their evaluations \citep{10.1073/pnas.2313790120}. Furthermore, the phenomenon of "hallucination," where LLMs produce plausible but factually incorrect information, poses a risk in academic contexts \citep{10.1177/05694345231218454}.

The role of LLMs in answering questions and generating hypotheses also merits attention. Their ability to provide detailed responses to complex queries has been leveraged in various educational settings, enhancing learning experiences and facilitating knowledge acquisition \citep{10.1088/1361-6404/ad1420}. However, the tendency of LLMs to produce verbose outputs can sometimes obscure the clarity of their answers, necessitating careful prompt engineering \citep{10.1088/1361-6552/ad1fa2}. In the context of academic research, LLMs can assist in generating hypotheses and guiding exploratory studies, contributing to the advancement of knowledge in various fields \citep{10.31235/osf.io/j2u9x}.

Despite the promising applications of LLMs in academic evaluation and research, it is crucial to establish ethical guidelines and best practices for their use. The potential for misuse, such as generating misleading information or facilitating academic dishonesty, necessitates careful consideration of the implications of LLM deployment in educational and research contexts \citep{10.31219/osf.io/q9v8f}.

\section{Methodology and Evaluation}
\subsection{LLM Evaluation of Titles and Abstracts}
\begin{figure*}[ht]
  \centering
  \includegraphics[width=\textwidth]{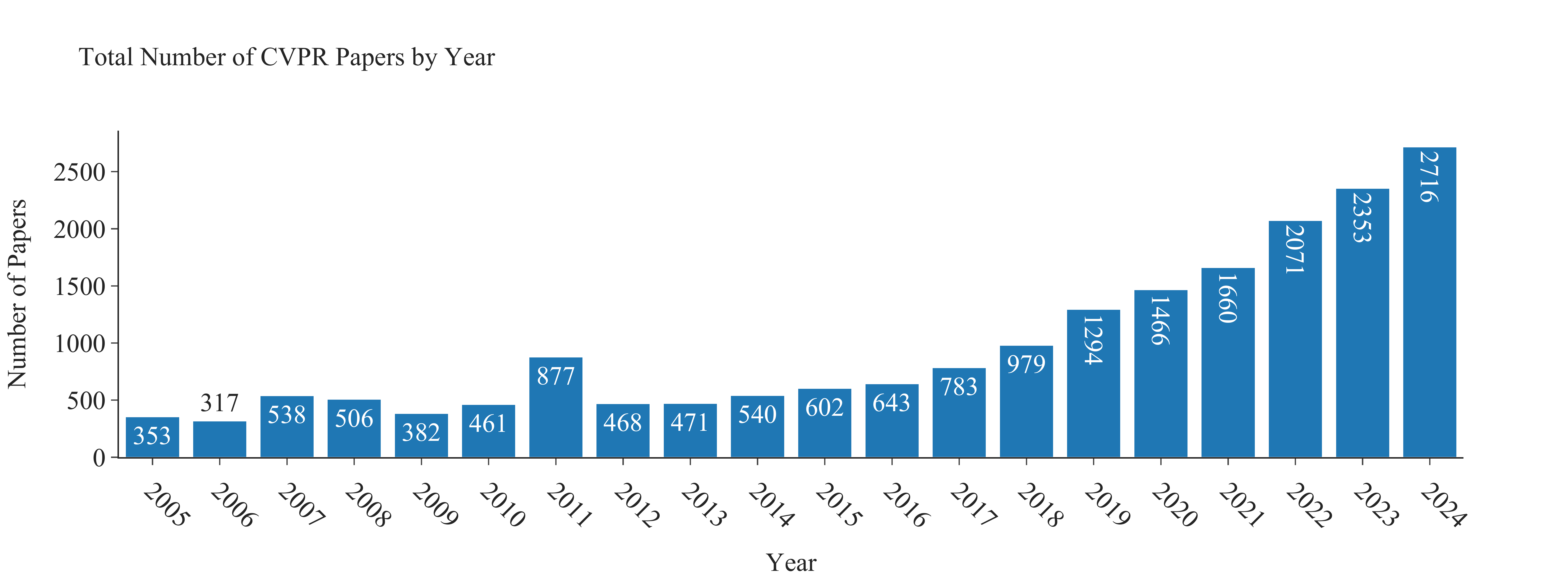}
  \caption{Total number of CVPR papers present in database for each year from 2005 to 2024.}
  \label{fig:total_papers}
\end{figure*}

We employ three large language models to evaluate the title and abstracts of CVPR papers from 2005 to 2024: GPT-4o-2024-05-13, gpt-4o-mini-2024-07-18, and claude-3-5-sonnet-20240620. The following information is extracted from online portals and stored in a database for each paper: Publication year (2005-2024), Title, Authors, Abstract. For each paper, the citation count from Semantic Scholar API is also queried on July 20th 2024, and stored alongside the other metadata. The total number of papers per year is shown in \autoref{fig:total_papers}.

Each LLM model is tasked with assigning a Likert score of 0-10 for how well the paper aligns with the principles of Sutton's "bitter lesson." We use the Chain-of-Thought Prompting technique with Magentic library to interface with the models and collect their responses in a structured format for analysis \citep{jack_collins_2024_13337526}. The prompts used in this study are included in the appendix for reproducibility.

We define five dimensions for "bitter lesson" alignment:

\begin{enumerate}
  \item \textbf{Learning Over Engineering}: To what extent does the idea prioritize leveraging computation through data-driven learning and statistical methods over relying on human-engineered knowledge, heuristics, and domain expertise?
  \item \textbf{Search over Heuristics}: To what degree does the idea emphasize leveraging computation through search algorithms and optimization techniques rather than depending on human-designed heuristics and problem-specific strategies?
  \item \textbf{Scalability with Computation}: To what extent is the idea based on methods that can continuously scale and improve performance as the available computational resources increase?
  \item \textbf{Generality over Specificity}: To what degree does the approach emphasize general, flexible, and adaptable methods that can learn and capture arbitrary complexity from data rather than attempting to build in complex and detailed models of the world through manual engineering and domain-specific knowledge?
  \item \textbf{Favoring Fundamental Principles}: To what extent does the approach adhere to fundamental principles of computation, mathematics, and information theory rather than focusing on emulating the specific details of human cognition or biological intelligence?
\end{enumerate}

The prompts were designed to capture the essence of each "bitter lesson" dimension concisely and objectively. To anchor the ratings, we provide examples for the 0, 5, and 10 points on each dimension, clarifying the criteria and ensuring consistent evaluations. The prompts are formatted consistently to facilitate easy processing and understanding by the models.

Given the vast number of publications, our study focuses on a representative random sample of 200 papers from each year of CVPR proceedings. We define the overall alignment score for each paper as the sum of scores across five dimensions. In the absence of human-evaluated ground truth, we employ multiple inter-rater reliability measures to assess the consistency of ratings between different models.

\subsection{Inter-rater Reliability Measures}

\textbf{Intraclass Correlation Coefficient (ICC):} We utilize ICC to quantify the degree of agreement among the models' ratings. ICC is particularly suitable for assessing reliability when multiple raters evaluate the same set of items. We specifically employ the two-way random effects model (ICC(2,k)) to account for both rater and subject effects.

\textbf{Krippendorff's Alpha:} To complement ICC, we also calculate Krippendorff's Alpha, a versatile reliability coefficient that can handle various types of data (nominal, ordinal, interval, ratio) and is robust to missing data. This measure provides an additional perspective on the inter-rater agreement, especially useful when dealing with potential variations in rating scales or missing evaluations.

\subsection{Regression Analysis}

To investigate the relationship between alignment scores and paper impact, we conduct regression analysis using citation count as a proxy for influence. To control for the year of publication and account for potential temporal effects, we implement yearly stratification in our regression model. This approach allows us to isolate the impact of alignment while considering the varying citation patterns across different publication years.

\begin{figure}[ht]
  \centering
  \includegraphics[width=\columnwidth]{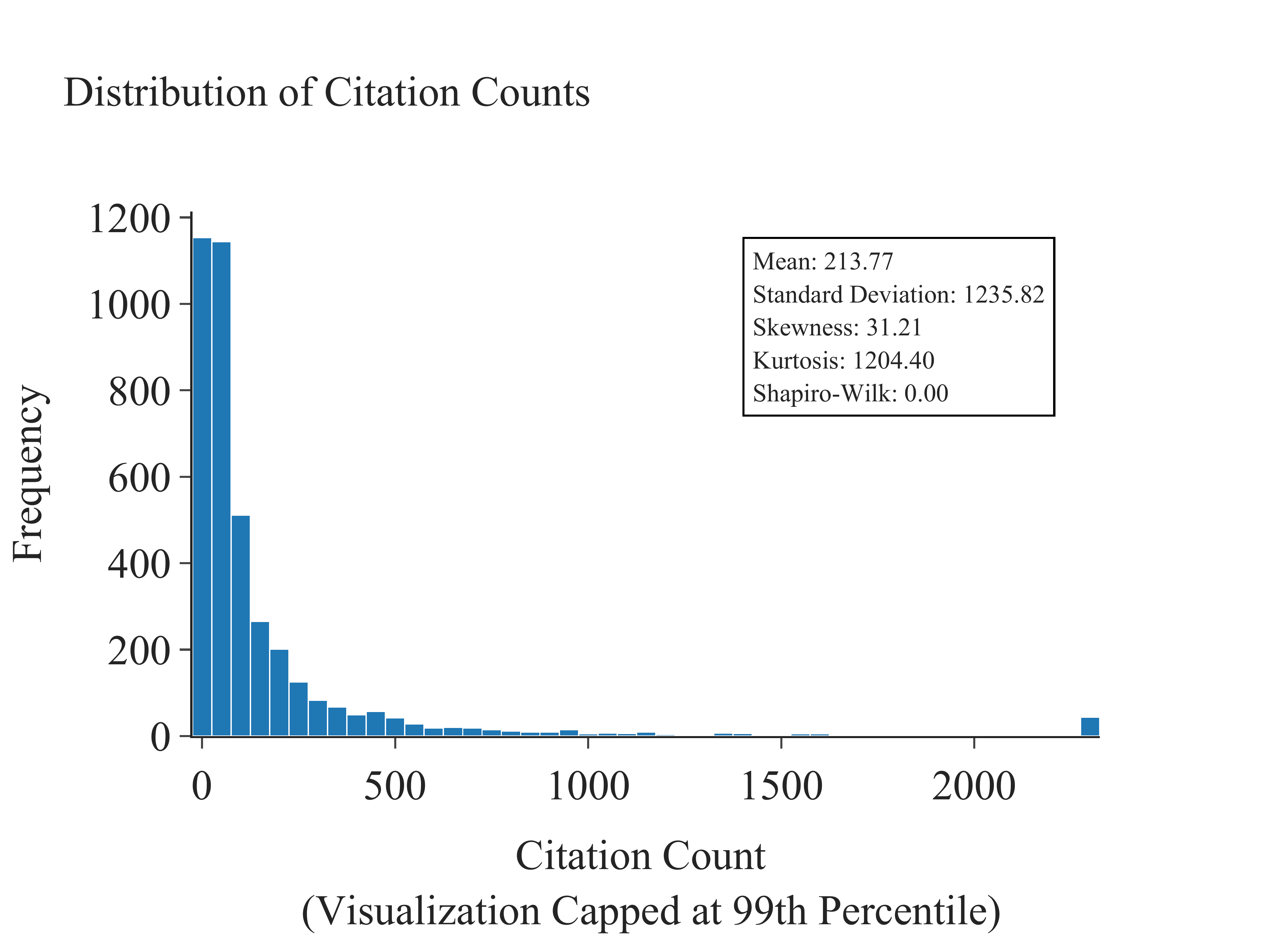}
  \includegraphics[width=\columnwidth]{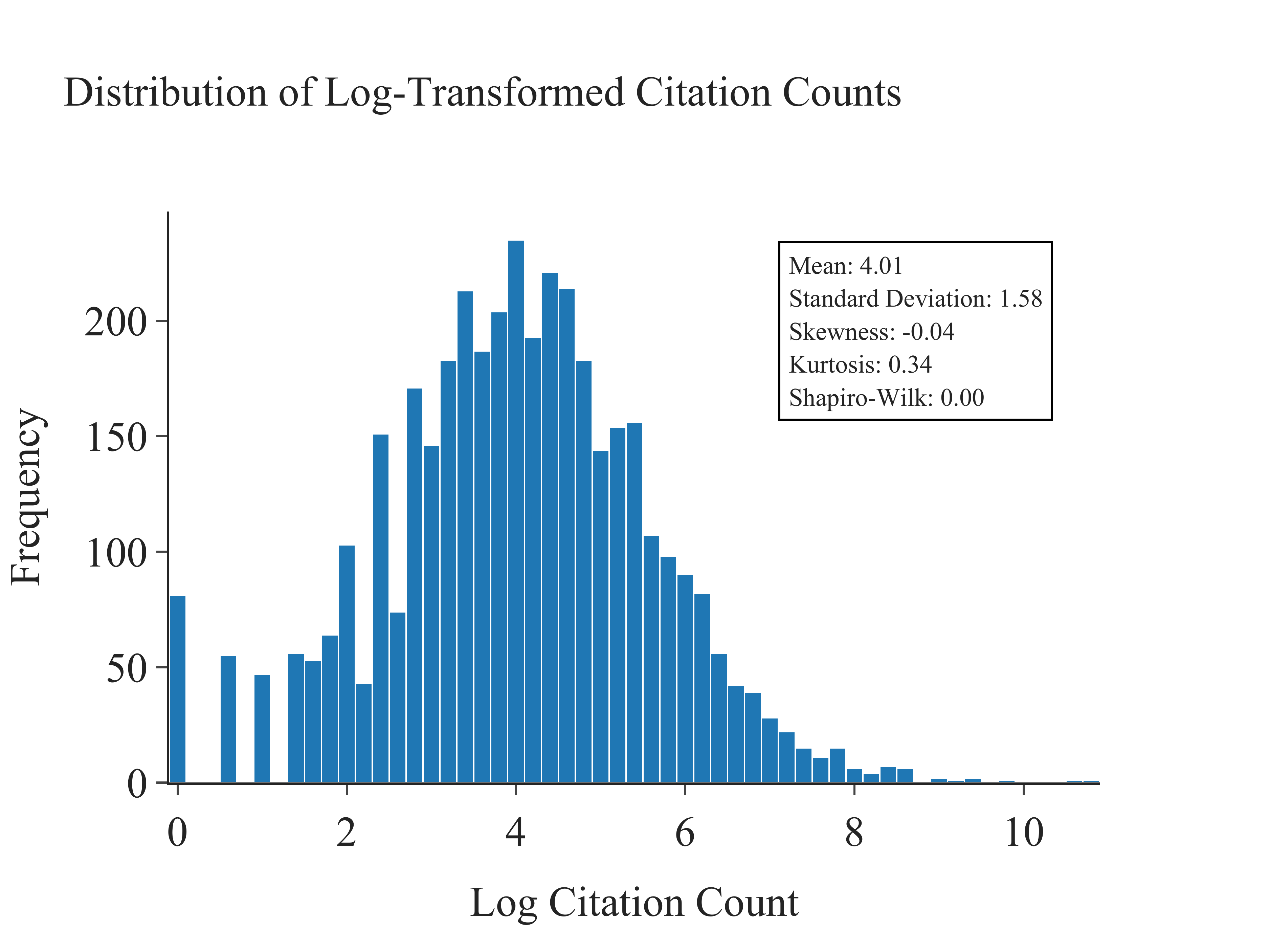}
  \caption{Distribution of citation counts and log-transformed citation counts for CVPR papers from 2005 to 2024 present in the database.}
  \label{fig:citation_distribution}
\end{figure}

To address the typically right-skewed distribution of citation counts \autoref{fig:citation_distribution}, we apply a logarithmic transformation to the data. This transformation serves multiple purposes in our analysis. First, it reduces skewness, resulting in a more symmetric distribution that better approximates normality—a key assumption in many statistical models. Second, it stabilizes variance across the range of data, mitigating the heteroscedasticity often observed in citation count data where variance tends to increase with the mean. Finally, the log transformation linearizes potentially multiplicative relationships, converting them to additive ones. This facilitates more accurate modeling using linear regression techniques, particularly when the effect of predictors on citation counts is expected to be multiplicative rather than additive. By employing this transformation, we enhance the robustness of our statistical analyses and ensure they are better suited to the inherent characteristics of citation data in academic literature.

The results of the analysis are presented in the following section.

\section{Results}

\subsection{Inter-rater Reliability}

\begin{figure*}[ht]
  \centering
  \includegraphics[width=0.49\textwidth]{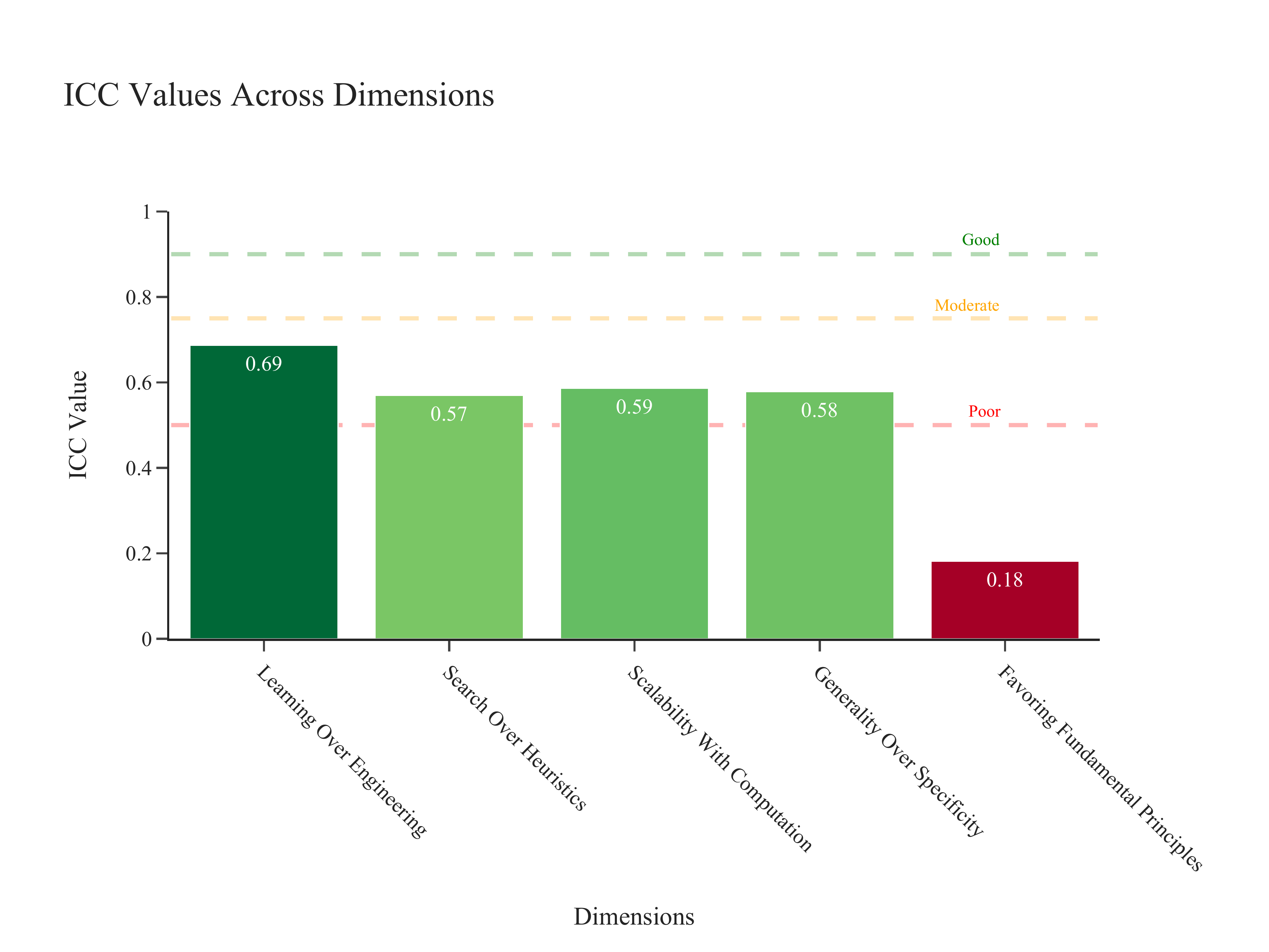}
  \hfill
  \includegraphics[width=0.49\textwidth]{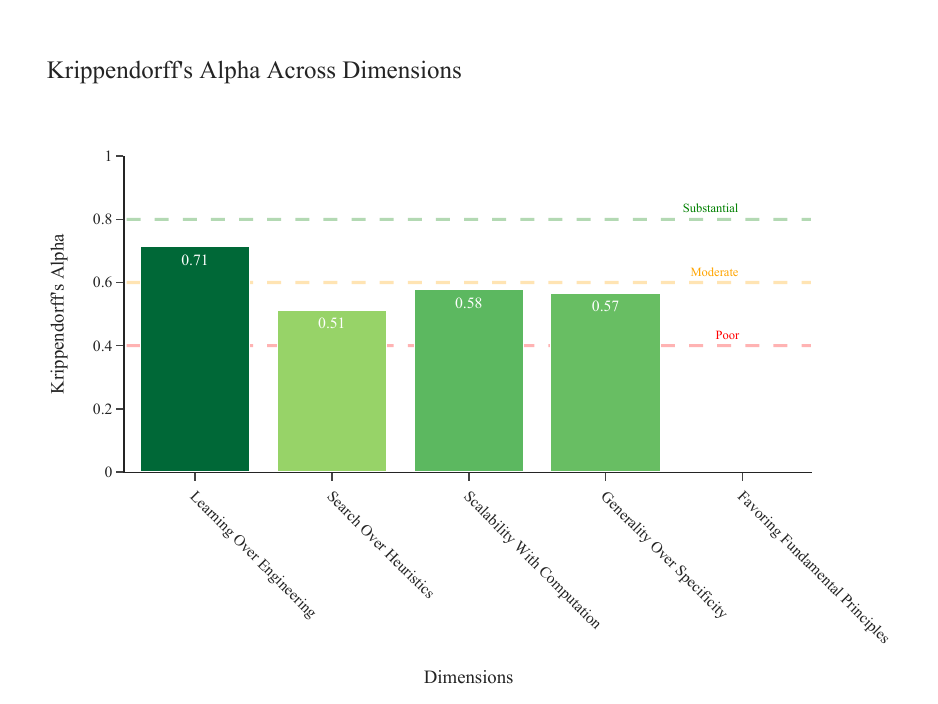}
  \caption{Comparison of ICC and Krippendorff's alpha values across the five dimensions of "bitter lesson" alignment for the three language models used in the study.}
  \label{fig:inter_rater_combined}
\end{figure*}

\autoref{fig:inter_rater_combined} presents the inter-rater reliability scores for the five dimensions of "bitter lesson" alignment across the three models employed in this study. The horizontal dashed lines indicate common thresholds for interpreting these measures, with the color and label denoting the qualitative interpretation. The bar colors reflect the relative strength of each dimension. In the Krippendorff's alpha graph, the bar for the "Favoring Fundamental Principles" dimension is not visible due to its near-zero score.

The models demonstrate consistently strong agreement on all dimensions except "Favoring Fundamental Principles," as evidenced by ICC values above 0.5 and Krippendorff's alpha scores exceeding 0.4 on the remaining dimensions. The poor performance on "Favoring Fundamental Principles" may be attributed to the high adherence to this principle in papers published since 2005. Among the dimensions, "Learning Over Engineering" exhibits the highest ICC and Krippendorff's alpha scores, indicating the models' reliable evaluation of paper alignment based on the provided prompts and rating criteria.

Although perfect agreement is not achieved, the inter-reliability measures fall within or above common thresholds for "good" reliability, validating the use of AI models for prompt-based research paper evaluation. It is important to acknowledge that even with domain expert human raters, perfect agreement is seldom attained due to the complexities of research evaluation. The high reliability scores obtained in this study demonstrate the models' consistency in their assessments, providing a reliable foundation for further analysis. For more information on the challenges and limitations of inter-rater reliability measures in human evaluations of the NeurIPS conference, readers may refer to \cite{Beygelzimer2023HasTM} and \cite{Cortes2021InconsistencyIC}.

\subsection{Regression Analysis}

\begin{table*}[h!]
  \centering
  \caption{Regression analysis results for the relationship between "bitter lesson" alignment scores and citation impact, stratified by year.}
  \label{tab:regression_results}
  \begin{tabular}{rrrlllll}
    \hline
    Year & R-squared & N   & Learning  & Search    & Scalability & Generality & Principles \\
    \hline
    2005 & 0.027     & 199 & -0.220    & 0.104     & 0.139       & 0.272      & -0.171     \\
    2006 & 0.076     & 200 & 0.016     & -0.042    & 0.388*      & 0.199      & -0.171     \\
    2007 & 0.035     & 200 & -0.087    & 0.117     & 0.350*      & -0.006     & -0.318*    \\
    2008 & 0.078     & 200 & -0.009    & 0.096     & 0.465***    & -0.026     & -0.463***  \\
    2009 & 0.085     & 199 & -0.073    & 0.136     & 0.104       & 0.378*     & -0.631***  \\
    2010 & 0.074     & 200 & 0.121     & -0.129    & 0.218       & 0.016      & -0.471**   \\
    2011 & 0.076     & 200 & 0.208     & -0.036    & 0.318**     & -0.284     & -0.423**   \\
    2012 & 0.094     & 200 & 0.195     & 0.077     & 0.428**     & -0.110     & -0.517**   \\
    2013 & 0.085     & 200 & 0.395***  & -0.112    & 0.013       & -0.119     & -0.279     \\
    2014 & 0.119     & 200 & 0.408***  & -0.085    & 0.308*      & -0.348*    & -0.266     \\
    2015 & 0.264     & 200 & 0.515***  & -0.145    & 0.417**     & -0.236     & -0.122     \\
    2016 & 0.306     & 200 & 0.637***  & -0.300**  & 0.517***    & -0.325     & -0.372*    \\
    2017 & 0.313     & 200 & 0.418***  & -0.353**  & 0.751***    & -0.004     & -0.508**   \\
    2018 & 0.172     & 200 & 0.291*    & -0.322*   & 0.418**     & 0.156      & -0.436**   \\
    2019 & 0.111     & 200 & 0.573**   & -0.439**  & 0.229       & -0.099     & -0.257     \\
    2020 & 0.120     & 200 & 0.315     & -0.411*** & 0.179       & 0.229      & 0.010      \\
    2021 & 0.090     & 200 & 0.269*    & -0.381*** & 0.253       & -0.072     & -0.265*    \\
    2022 & 0.136     & 200 & 0.618***  & -0.137    & 0.110       & -0.118     & -0.257     \\
    2023 & 0.123     & 200 & 0.107     & -0.009    & 0.664***    & -0.078     & -0.132     \\
    2024 & 0.178     & 171 & -0.619*** & 0.314     & 0.808***    & 0.282      & -0.020     \\
    \hline
  \end{tabular}

  \caption*{\raggedleft\footnotesize{*** indicates significance at the 1\% level, ** indicates significance at the 5\% level, and * indicates significance at the 10\% level.}}
\end{table*}

\autoref{tab:regression_results} presents the results of the regression analysis for each dimension of "bitter lesson" alignment scores against citation impact, stratified by year of publication. The R-squared values, ranging from 0.027 to 0.306, indicate that 2.7-30.6\% of the variation in citation impact can be explained by alignment to "bitter lessons" dimensions. It is crucial to interpret the coefficients for each dimension as multiplicative effects, as the log transform of citation counts is used as the dependent variable.

In the context of this regression analysis, a multiplicative effect implies that a one-unit change in the alignment score for a particular dimension leads to a proportional change in the original scale of the citation count. For instance, if the regression coefficient for the "Scalability" dimension is 0.5, a one-unit increase in the "Scalability" alignment score would be associated with a multiplicative effect of approximately $\exp(0.5) \approx 1.65$ on the original citation count. In other words, if a paper's "Scalability" alignment score increases by one unit, its citation count would be expected to increase by a factor of 1.65, holding all other variables constant.

The statistical significance of the regression coefficients is denoted using *, **, and *** to represent the 10\%, 5\%, and 1\% significance levels, respectively. Several dimensions, such as "Scalability" and "Learning over engineering," exhibit statistically significant relationships with citation impact across multiple years. However, given the high degree of correlation between the dimensions, the significance and coefficients in the regression model should be interpreted with caution.

These findings suggest that adherence to the principles outlined in the "bitter lesson" dimensions, particularly "Scalability" and "Learning over engineering," may have a positive influence on a paper's citation impact. The multiplicative nature of the coefficients highlights the potential for substantial increases in citation counts as alignment scores improve. Nevertheless, the presence of correlations among the dimensions necessitates a cautious interpretation of the individual coefficients and their statistical significance.

\begin{table*}[h!]
  \centering
  \caption{Regression analysis results for the relationship between overall "bitter lesson" alignment scores and citation impact,
    stratified by year.}
  \label{tab:overall_regression_result}
  \begin{tabular}{rrrrrl}
    \hline
    Year & R-squared & N   & F-statistic & Prob (F-statistic) & Overall Alignment Score \\
    \hline
    2005 & 0.007     & 199 & 1.409       & 0.237              & 0.029 [-0.019, 0.076]   \\
    2006 & 0.050     & 200 & 10.335      & 0.002              & 0.083*** [0.032, 0.134] \\
    2007 & 0.003     & 200 & 0.554       & 0.457              & 0.019 [-0.031, 0.068]   \\
    2008 & 0.010     & 200 & 1.993       & 0.160              & 0.031 [-0.012, 0.075]   \\
    2009 & 0.015     & 199 & 2.998       & 0.085              & 0.045* [-0.006, 0.097]  \\
    2010 & 0.000     & 200 & 0.033       & 0.856              & 0.005 [-0.049, 0.059]   \\
    2011 & 0.000     & 200 & 0.000       & 0.993              & -0.000 [-0.051, 0.051]  \\
    2012 & 0.024     & 200 & 4.898       & 0.028              & 0.057** [0.006, 0.109]  \\
    2013 & 0.005     & 200 & 0.944       & 0.333              & 0.022 [-0.023, 0.067]   \\
    2014 & 0.030     & 200 & 6.023       & 0.015              & 0.056** [0.011, 0.101]  \\
    2015 & 0.170     & 200 & 40.618      & 0.000              & 0.141*** [0.097, 0.184] \\
    2016 & 0.128     & 200 & 29.114      & 0.000              & 0.129*** [0.082, 0.176] \\
    2017 & 0.133     & 200 & 30.338      & 0.000              & 0.182*** [0.117, 0.248] \\
    2018 & 0.066     & 200 & 13.996      & 0.000              & 0.098*** [0.047, 0.150] \\
    2019 & 0.021     & 200 & 4.241       & 0.041              & 0.061** [0.003, 0.119]  \\
    2020 & 0.040     & 200 & 8.325       & 0.004              & 0.079*** [0.025, 0.133] \\
    2021 & 0.002     & 200 & 0.407       & 0.524              & -0.017 [-0.068, 0.035]  \\
    2022 & 0.062     & 200 & 13.054      & 0.000              & 0.097*** [0.044, 0.149] \\
    2023 & 0.063     & 200 & 13.416      & 0.000              & 0.099*** [0.046, 0.153] \\
    2024 & 0.092     & 171 & 17.040      & 0.000              & 0.127*** [0.066, 0.188] \\
    \hline
  \end{tabular}
  \caption*{\raggedleft\footnotesize{*** indicates significance at the 1\% level, ** indicates significance at the 5\% level, and * indicates significance at the 10\% level.}}
\end{table*}

\autoref{tab:overall_regression_result} shows the results of regressing citation counts on the overall "bitter lesson" alignment score for each year between 2005 and 2024. Several key trends emerge. First, the R-squared values, which indicate the proportion of variance in citation counts explained by the alignment scores, are quite low for most years (generally less than 5\%). However, they increase substantially starting in 2015, reaching over 15\% in some later years. This suggests that alignment with the "bitter lessons" became more predictive of citation impact over time. This time period is of special interest as it coincides with the emergence of deep learning, and a shift towards the principles of scalability and learning from data that are emphasized in the "bitter lessons."
Second, the overall alignment scores exhibit a statistically significant positive relationship with citations in many individual years, most prominently after 2011. The coefficients tend to be largest in later years as well. This indicates that as deep learning became more established, papers more closely adhering to principles like scalability and learning from data received more citations on average. The results suggest that the "bitter lessons" have become increasingly important in the field of computer vision, aligning with the broader trend towards data-driven methods and scalable algorithms in machine learning research.

\subsection{Trends in "Bitter Lesson" Alignment}

\begin{figure*}[h!]
  \centering
  \includegraphics[width=\textwidth]{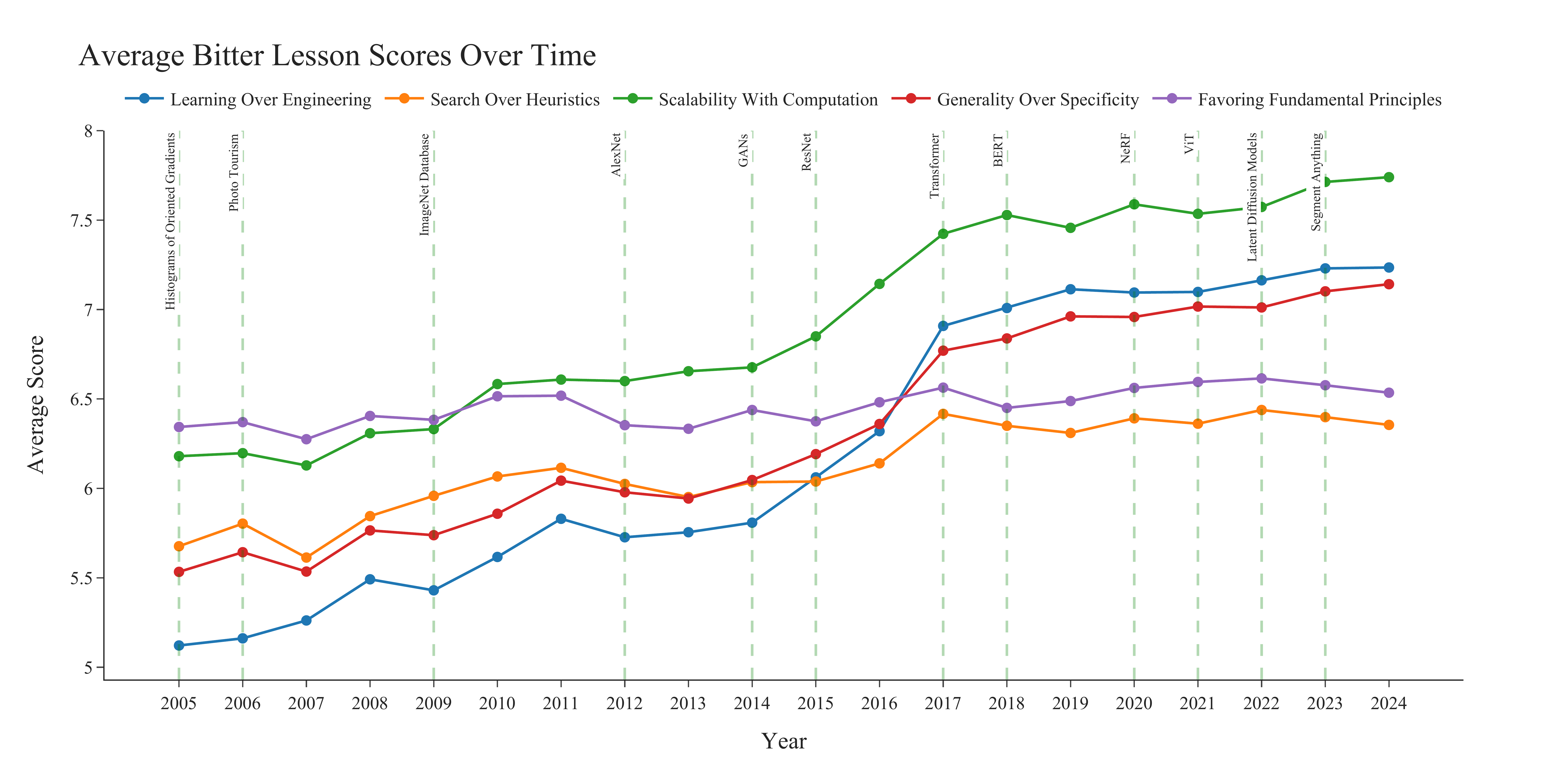}
  \caption{Line plot showing the average alignment scores across years for CVPR papers from 2005 to 2024.}
  \label{fig:line_plot_score_evaluation}
\end{figure*}

\autoref{fig:line_plot_score_evaluation} presents the average alignment scores for each "bitter lesson" dimension across the years 2005-2024. The vertical lines, which depict the publication of influential papers in machine learning (not necessarily computer vision), serve as a guide to understanding the overall evolution of the field. The averages are calculated across all papers and all language models (LLMs) employed in the study. The plot reveals several notable trends in the alignment of CVPR papers with the principles of the "bitter lesson." Notably, the dimensions of "Scalability with Computation" and "Learning Over Engineering" exhibit a consistent upward trend over the years, indicating a growing emphasis on scalable algorithms and data-driven learning methods in CVPR research. This trend aligns with the broader shift towards deep learning and foundation models in computer vision, emphasizing the importance of leveraging computation and large datasets for superior performance.

The period from 2015 to 2020 witnesses a particularly sharp rise in the average scores for these dimensions, coinciding with major advances in deep learning, such as the application of convolutional neural networks to computer vision tasks. Interestingly, this time frame corresponds to the period in which the regression analysis finds the highest predictive power of alignment scores on citation counts. This finding suggests that the increasing alignment of CVPR papers with the principles of scalability and learning-oriented approaches during this period has a significant impact on their academic influence, as measured by citation metrics.

The observed trends in the alignment scores highlight the evolving landscape of computer vision research, with a growing emphasis on leveraging the power of computation and data-driven learning techniques. The coincidence of these trends with the increased predictive power of alignment scores on citation counts underscores the importance of adhering to the principles of the "bitter lesson" for achieving impactful research outcomes in the field of computer vision.

\section{Conclusion}

Our study examined the alignment of CVPR research with Rich Sutton's "The Bitter Lesson" over twenty years, leveraging large language models to analyze trends. The findings reveal a consistent increase in the adoption of general-purpose learning algorithms and scalability with computational resources, reflecting a strong adherence to the core principles of the "bitter lesson." These trends underscore the machine learning community's preference for data-driven and computation-heavy approaches over manual engineering and domain-specific knowledge.

However, the dimension of "Search over Heuristics" has not experienced a similar upward trajectory, indicating limited integration of search-based methodologies within the field. This stagnation contrasts with recent advancements in inference-time scaling, exemplified by OpenAI's o1 models, which emphasize the importance of test-time compute in overcoming diminishing returns. The o1 models' ability to simulate various strategies and scenarios during inference, similar to AlphaGo’s Monte Carlo Tree Search (MCTS), marks a key departure from earlier approaches that relied heavily on large pre-trained models.

The paradigm shift towards scaling inference time, driven by the development of larger and more complex models, has the potential to emulate search-like processes. As computational capabilities continue to expand, it is plausible that future research may increasingly incorporate search techniques, thereby enhancing alignment with this dimension of the "bitter lesson." The dynamic resource allocation in o1 models, which adjusts computational resources based on task complexity, further underscores the potential for integrating search methodologies.

Overall, our findings highlight the continued relevance of the "bitter lesson" in shaping the trajectory of computer vision research. By emphasizing generality and scalability, the field is well-positioned to leverage emerging computational advancements. Future work should explore the integration of search methodologies and assess their impact on research impact and innovation within computer vision, particularly in light of recent breakthroughs in inference-time scaling.

\section*{Limitations}

This study, while providing valuable insights into the evolution of computer vision research, has several limitations that should be acknowledged. Firstly, our reliance on large language models (LLMs) for evaluating research abstracts, while innovative, introduces potential biases inherent to these models. The LLMs' understanding and interpretation of complex scientific concepts may not always align perfectly with human expert judgment.

Secondly, the absence of human expert evaluation as a ground truth is a significant limitation. Collecting such human evaluations presents considerable challenges, as it would require a diverse panel of researchers from various subfields of the computer vision community. The interdisciplinary nature of modern computer vision research necessitates expertise in areas ranging from traditional image processing to deep learning, computer graphics, and even cognitive science. Assembling such a panel and achieving consensus on the evaluation criteria would be a formidable task, both in terms of logistics and resources.

Furthermore, our analysis is limited to the information contained in titles and abstracts. While these elements provide a concise summary of research, they may not capture the full depth and nuance of the methodologies and findings presented in the full papers. This limitation could potentially lead to oversimplification of complex research ideas.

Lastly, while our study spans two decades of CVPR proceedings, it does not account for research published in other venues or unpublished work that may have influenced the field. This focus on a single conference, albeit a prestigious one, may not provide a complete picture of the entire computer vision research landscape.

Despite these limitations, we believe our study provides valuable insights into broad trends in computer vision research and its alignment with the principles of the "bitter lesson." Future work could address these limitations by incorporating human expert evaluations, analyzing full paper contents, and expanding the scope to include a wider range of publication venues.

\section*{Ethics Statement}

This study adheres to the ACL Ethics Policy. Our use of large language models (LLMs) for analyzing trends in academic literature raises important ethical considerations. We acknowledge that LLMs may introduce biases when used for direct evaluation of academic work. However, our study focuses solely on using LLMs to analyze broad trends rather than to assess individual papers' quality or merit. We have addressed the challenges and potential biases of LLM use for evaluation in our background section, emphasizing the need for careful interpretation of results.

All data were collected in accordance with applicable privacy and intellectual property laws. The titles and abstracts of CVPR papers were collected from the conference website, which allows for such collection and analysis under standard terms of use. Citation counts were collected from Semantic Scholar, which also permits such collection and analysis under its standard terms of use. No personally identifiable information was collected from human subjects.

Our methodology aims to minimize risks by using multiple models and focusing on aggregate trends rather than individual assessments. No crowd workers or annotators were involved in the data collection process described in the paper. We believe this approach provides valuable insights into the evolution of computer vision research while maintaining ethical standards in AI-assisted academic analysis.

\bibliography{anthology,custom}
\bibliographystyle{acl_natbib}

\clearpage
\onecolumn

\appendix
\section{Prompt and Example Usage}
\label{sec:appendix}

\setminted{
  frame=lines,
  framesep=2mm,
  baselinestretch=1.2,
  fontsize=\footnotesize,
  breaklines=true,
  breakanywhere=true
}

\begin{minted}{python}
from magentic import prompt
from pydantic import BaseModel, Field


class Score(BaseModel):
    explanation: str = Field(description="An explanation for the given score")
    score: int = Field(
        description="A score from 0 to 10",
        ge=0,
        le=10,
    )


class BitterLessonScores(BaseModel):
    learning_over_engineering_score: Score = Field(
        description="**Learning Over Engineering**: To what extent does the idea prioritize leveraging computation through data-driven learning and statistical methods (e.g., machine learning, deep learning, neural networks, probabilistic models, unsupervised learning, supervised learning, reinforcement learning, generative models, discriminative models, ensemble methods, online learning, active learning, semi-supervised learning) over relying on human-engineered knowledge, heuristics, and domain expertise (e.g., hand-crafted features, rule-based systems, expert systems, symbolic AI, knowledge representation, logic programming, constraint satisfaction)?\n\nPlease rate on a scale from 0 to 10, where:\n0 = Completely relies on human engineering, 5 = Equal emphasis on learning and engineering, 10 = Completely prioritizes learning from data",
    )
    search_over_heuristics_score: Score = Field(
        description="**Search over Heuristics**: To what degree does the idea emphasize leveraging computation through search algorithms (e.g., gradient descent, stochastic gradient descent, evolutionary algorithms, genetic algorithms, simulated annealing, Monte Carlo methods, Markov chain Monte Carlo, beam search, branch and bound, A* search, heuristic search) and optimization techniques (e.g., convex optimization, stochastic optimization, combinatorial optimization, integer programming, quadratic programming, linear programming, non-linear optimization, multi-objective optimization), rather than depending on human-designed heuristics and problem-specific strategies (e.g., hand-tuned parameters, domain-specific rules, expert knowledge, case-based reasoning, heuristic functions)?\n\nPlease rate on a scale from 0 to 10, where:\n0 = Completely relies on human-designed heuristics, 5 = Equal emphasis on search and heuristics, 10 = Completely prioritizes search and optimization",
    )
    scalability_with_computation_score: Score = Field(
        description="**Scalability with Computation**:To what extent is the idea based on methods that can continuously scale and improve performance as the available computational resources (e.g., processing power, memory, storage, data, distributed computing, cloud computing, GPU acceleration, TPU acceleration, high-performance computing, edge computing, quantum computing) increase, taking full advantage of the exponential growth in computing capabilities (e.g., Moore's Law, Dennard scaling, Amdahl's Law, Gustafson's Law)?\n\nPlease rate on a scale from 0 to 10, where:\n0 = Does not scale with computation at all, 5 = Scales moderately with computation, 10 = Scales exceptionally well with computation",
    )
    generality_over_specificity_score: Score = Field(
        description="**Generality over Specificity**:To what degree does the approach emphasize general, flexible, and adaptable methods that can learn and capture arbitrary complexity from data (e.g., deep learning, transfer learning, meta-learning, representation learning, multi-task learning, few-shot learning, zero-shot learning, self-supervised learning, unsupervised pre-training, domain adaptation, continual learning, lifelong learning, incremental learning) rather than attempting to build in complex and detailed models of the world through manual engineering and domain-specific knowledge (e.g., hand-designed features, domain-specific ontologies, knowledge graphs, expert systems, rule-based systems, symbolic representations, logic-based representations)?\n\nPlease rate on a scale from 0 to 10, where:\n0 = Completely domain-specific and manually engineered, 5 = Balance of generality and specificity, 10 = Maximally general, flexible and adaptable",
    )
    favoring_fundamental_principles_score: Score = Field(
        description="**Favoring Fundamental Principles**: To what extent does the approach adhere to fundamental principles of computation, mathematics, and information theory (e.g., algorithmic efficiency, computational complexity, statistical learning theory, information entropy, Bayesian inference, Kolmogorov complexity, Occam's razor, Minimum Description Length, PAC learning, VC dimension, Rademacher complexity, concentration inequalities, regularization, sparsity, smoothness, stability, convergence, consistency) rather than focusing on emulating the specific details of human cognition or biological intelligence (e.g., neuroscience-inspired architectures, cognitive architectures, embodied cognition, situated cognition, enactivism, dynamical systems theory, ecological psychology)?\n\nPlease rate on a scale from 0 to 10, where:\n0 = Completely focused on emulating human/biological details, 5 = Equal focus on principles and human/biological details, 10 = Completely grounded in fundamental principles",
    )


@prompt(
    """
Title: {title}
Abstract: {abstract}

We want to evalute this abstract in terms of alignment with "The Bitter Lesson". The main idea of Rich Sutton's "The Bitter Lesson" is that the most effective AI approaches in the long run are those that leverage computation and general-purpose methods like search and learning, rather than human-designed systems that try to build in human knowledge. Evaluate the alignment of the abstract with the following principles, assigning a score from 0 to 10 for each.
"""
)
def evaluate_bitter_lesson_alignment(
    title: str, abstract: str
) -> BitterLessonScores: ...


## EXAMPLE USAGE

bitter_lesson_scores = evaluate_bitter_lesson_alignment(
    title="Attention Is All You Need",
    abstract="The dominant sequence transduction models are based on complex recurrent or convolutional neural networks in an encoder-decoder configuration. The best performing models also connect the encoder and decoder through an attention mechanism. We propose a new simple network architecture, the Transformer, based solely on attention mechanisms, dispensing with recurrence and convolutions entirely. Experiments on two machine translation tasks show these models to be superior in quality while being more parallelizable and requiring significantly less time to train. Our model achieves 28.4 BLEU on the WMT 2014 English-to-German translation task, improving over the existing best results, including ensembles by over 2 BLEU. On the WMT 2014 English-to-French translation task, our model establishes a new single-model state-of-the-art BLEU score of 41.8 after training for 3.5 days on eight GPUs, a small fraction of the training costs of the best models from the literature. We show that the Transformer generalizes well to other tasks by applying it successfully to English constituency parsing both with large and limited training data.",
)
print(bitter_lesson_scores.model_dump_json(indent=2))
# {
#     "learning_over_engineering_score": {
#         "explanation": "The abstract describes a model called the Transformer that is based solely on attention mechanisms, dispensing with recurrence and convolutions. This indicates a strong reliance on learning from data rather than on human-engineered features or domain-specific knowledge. The significant improvement in BLEU scores across multiple tasks further showcases the efficacy of data-driven learning methods.",
#         "score": 9,
#     },
#     "search_over_heuristics_score": {
#         "explanation": "The Transformer model prioritizes the use of attention mechanisms to learn representations from data, which can be considered a form of search over heuristics. The architecture allows for efficient computation and optimization during training, indicating a significant emphasis on leveraging search algorithms and optimization techniques rather than relying on human-designed heuristics.",
#         "score": 8,
#     },
#     "scalability_with_computation_score": {
#         "explanation": "The abstract highlights the Transformer model's parallelizability and reduced training time, which suggests that the model can scale effectively with increased computational resources. The use of GPUs to achieve state-of-the-art performance in a relatively short training time further indicates that the model benefits significantly from additional computational power.",
#         "score": 9,
#     },
#     "generality_over_specificity_score": {
#         "explanation": "The abstract demonstrates the generality of the Transformer model by applying it successfully to multiple tasks, including machine translation and English constituency parsing. The model's ability to generalize well to tasks with both large and limited training data suggests that it is highly adaptable and not limited to specific domains or tasks.",
#         "score": 9,
#     },
#     "favoring_fundamental_principles_score": {
#         "explanation": "The Transformer model is grounded in fundamental principles of computation and information theory, particularly through its use of attention mechanisms, which can be seen as an efficient way to handle sequence transductions. The focus on parallelizability and optimization also aligns with fundamental principles rather than attempting to emulate human cognition or biological processes.",
#         "score": 8,
#     },
# }
\end{minted}

\end{document}